\documentclass[final,review,authoryear,11pt]{elsarticle}
\usepackage{lipsum}
\makeatletter
\def\ps@pprintTitle{
 \let\@oddhead\@empty
 \let\@evenhead\@empty
 \def\@oddfoot{}%
 \let\@evenfoot\@oddfoot}
\makeatother

\usepackage{natbib}
\usepackage{amssymb}
\usepackage{multirow}
\usepackage{multicol}
\usepackage{booktabs}
\usepackage{makecell}
\usepackage{fullpage}
\usepackage{lscape}
\usepackage{tikz}
\usepackage{graphicx} 
\usepackage{amsmath}
\usepackage{amsthm}
\usepackage{float}
\usepackage{rotating}
\usepackage{todonotes}
\usepackage{graphicx}
\usepackage{caption}
\usepackage{textcomp}

\usepackage{xcolor,colortbl}
\usepackage{url}
\usepackage{enumitem}
\usepackage{xcolor}
\usepackage{tabu}
\usepackage{multirow}
\usepackage[font={footnotesize}]{caption}
\usepackage{setspace}
\usepackage{rotating}
\usepackage{subfig}
\usepackage{xcolor}
\usepackage{ctable }
\usepackage{algpseudocode}
\usepackage{algorithm}

\usepackage[font={footnotesize}]{caption}
\usepackage{setspace}
\usepackage{rotating}
\usepackage{subfig}
\usepackage{xcolor}
\usepackage{subfig}
\usepackage{hhline}
\usepackage{lscape}
\usepackage{multirow}
\usepackage[flushleft]{threeparttable}
\usepackage{tikz}
\def\checkmark{\tikz\fill[scale=0.4](0,.35) -- (.25,0) -- (1,.7) -- (.25,.15) -- cycle;} 
\usepackage[hmargin = 3cm]{geometry} 
\def\mc#1#2{\multicolumn{#1}{c}{#2}}
\def\mr[#1]#2#3{\multirowcell{#2}[#1]{#3}}

\begin{document}

\begin{frontmatter}

\title{{\LARGE A Deep Active Survival Analysis Approach for Precision Treatment Recommendations: Application of Prostate Cancer}}

\author {Milad Zafar Nezhad$^{a,*}$}
\ead{m.zafarnezhad@wayne.edu}
\author {Najibesadat Sadati$^{a}$}
\ead{n.sadati@wayne.edu}
\author {Kai Yang$^{a}$}
\ead{kai.yang@wayne.edu}
\author {Dongxiao Zhu$^{b}$}
\ead{dzhu@wayne.edu}

\address {$^{a}$Department of Industrial and Systems Engineering, Wayne State University, \\4815 Fourth Street, Detroit, MI, 48202, USA}
\address {$^{b}$Department of Computer Science, Wayne State University, \\5057 Woodward Ave, Detroit, MI, 48202, USA}
\cortext[mycorrespondingauthor]{Corresponding author}
\begin{abstract}
Survival analysis has been developed and applied in the number of areas including manufacturing, finance, economics and healthcare. In healthcare domain, usually clinical data are high-dimensional, sparse and complex and sometimes there exists few amount of time-to-event (labeled) instances. Therefore building an accurate survival model from electronic health records is challenging. With this motivation, we address this issue and provide a new survival analysis framework using deep learning and active learning with a novel sampling strategy. First, our approach provides better representation with lower dimensions from clinical features using labeled (time-to-event) and unlabeled (censored) instances and then actively trains the survival model by labeling the censored data using an oracle. As a clinical assistive tool, we introduce a simple effective treatment recommendation approach based on our survival model. In the experimental study, we apply our approach on SEER-Medicare data related to prostate cancer among African-Americans and white patients. The results indicate that our approach outperforms significantly than baseline models.            

\end{abstract}

\begin{keyword}
Survival analysis \sep deep learning \sep active learning \sep treatment recommendation \sep electronic health records \sep prostate cancer 
\end{keyword}

\end{frontmatter}

\section{Introduction}
Survival analysis has been applied in several real-world applications such as healthcare, manufacturing and engineering in order to model time until the occurrence of an future event of interest (e.g. biological death or mechanical failure) \citep{hosmer2011applied}. Censoring attribute of survival data makes survival analysis different from the other prediction approaches. One popular survival model is the Cox Proportional Hazards model (CPH) \citep{cox1992regression} which models the risk of an event happening based on linear combination of the covariates (risk factors). The major problem of Cox-based models is linear relationship assumption between covariates and the time of event occurrence. Hence, there have been developed several models to handle non-linear relationship in survival analysis like as survival neural network and survival random forest models.      

In the healthcare area, medical researchers apply survival analysis on EHRs to evaluate the significance of many risk factors in outcomes such as survival rates or cancer recurrence and subsequently recommend treatment schemes. There exist two specific challenges in survival analysis from EHRs: 1) Clinical data is usually high dimensional, sparse and time-dependent which in this case applying traditional survival approaches do not perform well enough to estimate the risk of a medical event, 2) In many health survival applications, labeled data (time-to-event instances) are small, time-consuming and expensive to collect. In this situation, it is hard to learn a survival model based on traditional approaches which able to predict the relative risk of patients precisely. 

To address the first challenge, recently, semi-supervised learning using deep feature representation has been applied in number of areas and could improve the performance of different machine learning tasks as well as survival analysis. In the other word, applying unsupervised learning using deep learning can reduce the complexity of raw data and provide robust features with lower dimensions. Using this represented features in the supervised learning algorithms (e.g. survival models) establishes a semi-supervised learning framework which achieve higher performance. 

To overcome the second challenge, active learning is well suited to get high accuracy when the labeled instances are small or labeling is expensive and time-consuming. Active learning approach from censored data has been rarely addressed in the literature. However it has been widely used in the other aspects of health informatics where the labeled data are scarce. 

In this research, first, we propose a novel survival analysis approach using deep learning and active learning termed DASA. Our method is capable to learn more accurate survival model using high dimensional and small size EHRs in comparison with some baseline survival approaches. Second, we introduce a personalized treatment recommendation approach based on our survival analysis model which can compare the relative risk (or survival time) associate with different treatment plans and assign better one. We evaluate our approach using SEER-Medicare dataset related to prostate cancer. We consider two racial subgroup of patients (African-American and whites) in our analysis and apply our model on each dataset separately.    

Our contributions in this research lie into three folds: 1) To best of our knowledge, we propose the first deep active survival analysis approach with promising performance, 2) In our active learning framework we develop a new sampling strategy specifically for survival analysis and 3) Our model with proposed treatment recommendation approach is highly potential to apply for evaluation of new treatment effect on new patients where the labeled data is scarce.

\section{Background}
In this section, we review some basic concepts and the approaches for modeling of survival analysis, active learning and deep learning. 

\subsection{Introduction to Survival Analysis}
Survival analysis is a kind of statistical modeling where the main goal is to analyze and model time until the occurrence of an event of interest, such as death in biological systems and failure in mechanical machines. The challenging characteristic of survival data is the fact that time-to-event of interest for many instances is unknown because the event might not have happened during the period of study or missing tracking occurred caused by other events. This concept is called censoring which makes the survival analysis different. The special case of censoring is when the observed survival time is less than or equal to the true event time called right-censoring, the main focus of our study.

Since the censored data is present in survival analysis, the standard statistical and machine learning approaches are not appropriate to analyze and predict time-to-event outcome because those approaches miss the censored/right-censored instances. Survival modeling provides different statistical approaches to analyze such censored data in many real-world applications. 

In survival analysis, a given instance $i$, represented by a triplet ($X_{i}$, $\delta_{i}$, $T_{i}$) where $X_{i}$ refers to the instance characteristics and $T{i}$ indicates time-to-event of the instance. If the event of interest is observed, $T{i}$ corresponds to the time between baseline time and the time of event happening, in this case $\delta_{i}$ = 1. If the instance event is not observed and its time to event is greater than the observation time, $T{i}$ corresponds to the time between baseline time and end of the observation, and the event indicator is $\delta_{i}$ = 0. The goal of survival analysis is to estimate the time to the event of interest ($T$) for a new instance $X_{j}$. 

Survival and hazard functions are the two main functions in survival modeling. The survival function indicates the probability that the time to the event of interest is not less than a determined time ($t$). This function ($S$) is denoted by following formula:
\begin{align}\label{equ1}
&S(t) = Pr (T>t)&
\end{align}

The initial value of survival function is 1 when $t=0$  and it monotonically decreases with $t$. The second function, hazard function indicates the rate of occurrence of the event at time $t$ given that no event occurred earlier. It describes the risk of failure (dying) changing over time. The hazard function (or hazard rate or failure rate) is defined as following:
\begin{align}\label{equ1}
&h(t) = \lim_{\delta(t) \rightarrow 0} \dfrac{Pr(t\leq T\leq t+\delta(t) | T\geq t)}{\delta(t)}&
\end{align} 

Survival and hazard function are  non-negative functions. While the survival function decreases over time, The shape of a hazard function can be in different forms: increasing, decreasing, constant, or U-shaped. 

There exist several models for survival analysis in the literature. Among all, Cox Proportional Hazards (CPH) model \citep{cox1992regression} is the most popular model for survival analysis. CPH estimates the hazard function $h(x)$ as a regression formulation:
\begin{align}\label{equ1}
&h(t,X_{i}) = h_{0} \ exp(X_{i}\beta)&
\end{align} 

where $h_{0}$ is the baseline hazard function which can be an arbitrary nonnegative function of time and $X_{i}$ refers to covariate vector
for instance $i$, and $\beta$ is the coefficient vector estimated after survival model training by maximizing the cox partial likelihood. Because the baseline hazard function $h_{0}(t)$ in CPH is not determined, we cannot use the standard likelihood function in training process \citep{cox1992regression}. The partial likelihood is the product of the probability of each instance $i$ at event time $T_{i}$ that the event has happened for that instance, over the summation of instances ($R_{j}$) probability who are still at risk in this time ($T_{i}$):
\begin{align}\label{equ1}
&L(\beta) = \prod_{i=,\delta_{i}=1} \dfrac{exp(X_{i}\beta)}{\sum_{j \in R_{j}}exp(X_{j}\beta)}&
\end{align} 

Since the censored instances exist in survival data, the standard evaluation metrics such as mean squared error and R-squared are not appropriate for evaluating the performance of survival analysis \citep{heagerty2005survival}. In survival analysis, the most popular evaluation metric is based on the relative risk of an event for different instances called concordance index or c-index. This measure is defined as following formula:
\begin{align}\label{equ1}
&\frac{1}{N} \sum_{i, \delta_{i}=1} \sum_{j, y_{i} < y_{j}} I[S(\hat{y_{i}} | X_{i}) < S(\hat{y_{j}} | X_{j})]&
\end{align} 

Where $N$ refers to the all comparable instance pairs and $S$ is the survival function. The main motivation for using c-index in survival analysis is originated from the fact that the medical doctors and researchers are often more interested in measuring the relative risk of a disease among patients with different risk factors, than the survival times of patients. 

In general, the survival analysis models can be divided into two main categories: 1) statistical methods including non-parametric, semi-parametric and parametric and 2) machine learning based methods such survival trees, bayesian methods, neural networks and random survival forests. Readers for more comprehensive review can refer to the recent review provided by \cite{wang2017machine}. 

\subsection{Introduction to Active Learning}
Active learning is a subfield of machine learning and statistical modeling. The goal of an active learner is the same as a passive
learner but the key idea behind active learning is that a machine learning algorithm can lead to better performance with fewer training labels if it can select the data for learning. An active learner chooses queries, usually in the form of unlabeled data instances to be labeled by an oracle which can be a human annotator. Active learning is very efficient in many data-driven applications, where there exist numerous unlabeled data but labels are rare, time-consuming, or expensive to be labeled \citep{settles2010active}.

Since large amounts of unlabeled data is nowadays often available and can be easily collected by automatic processes, active learning would be demanding in modern applications in order to reduce the cost of labeling. The active learning framework overcomes the challenge of insufficient labeled data by efficiently modeling the process of obtaining labels for unlabeled data. The advantage is that the active learner just requires to query the labels of just a few, carefully selected instances during the iterative process in order to achieve more accurate learner \citep{hsu2010algorithms}. 

There exist several approaches/scenarios in which active learners ask queries. The three main approaches widely used in the literature are \citep{settles2010active}: 1) membership query synthesis \citep{angluin1988queries}, 2) stream-based selective sampling \citep{atlas1990training}, and 3) pool-based sampling \citep{lewis1994sequential}. For all approaches, there are also several different query strategies that have been developed to decide which unlabeled instances should be selected. Among above three approaches, pool-based sampling is most popular in many real-world applications. This approach has been demonstrated in Figure \ref{fig.4.1}:

\begin{figure}[H]
	\centering
	\includegraphics[scale= 0.1]{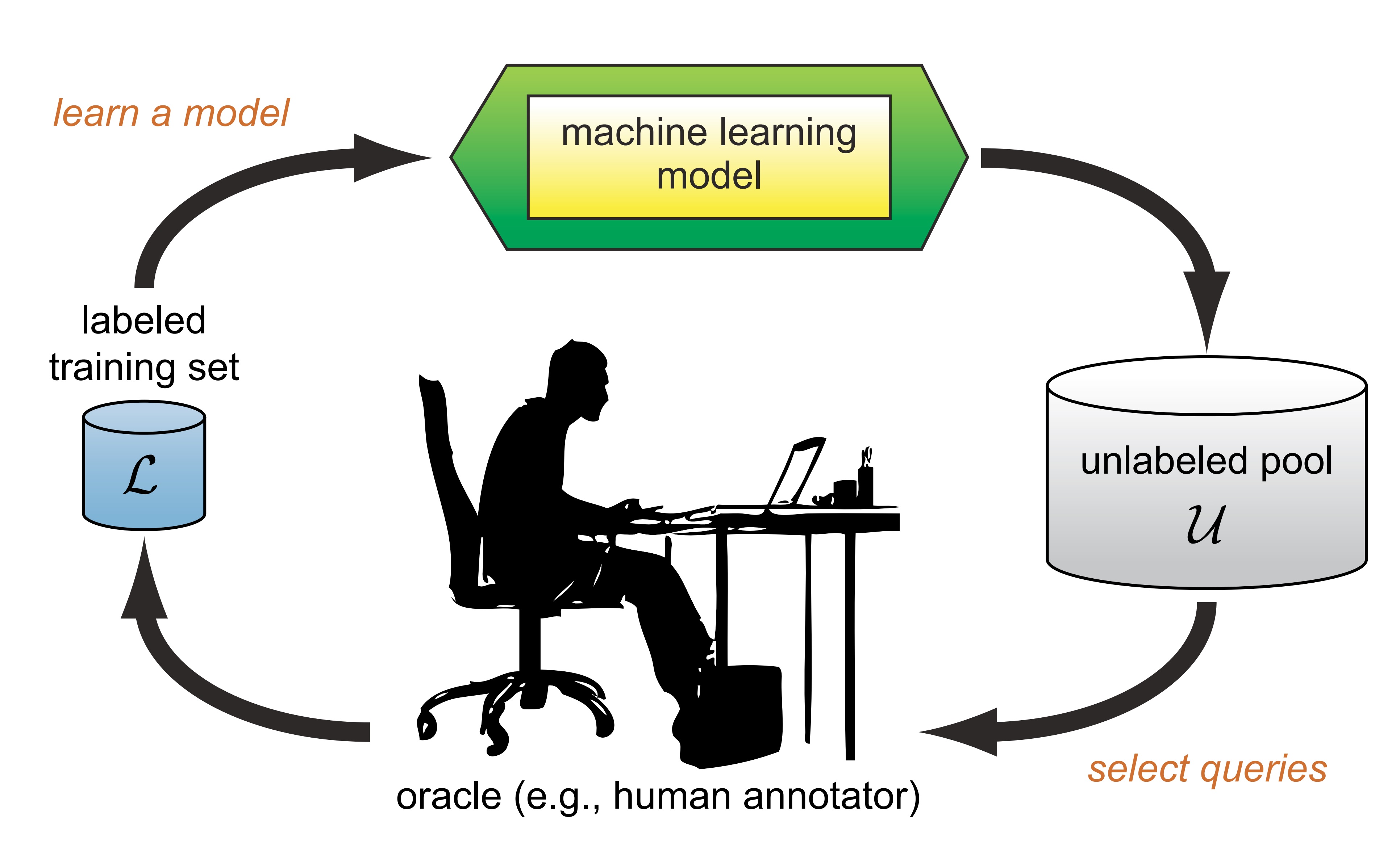} 
	\caption{The pool-based active learning approach \citep{settles2010active} }
	\label{fig.4.1}
\end{figure}

According to Figure \ref{fig.4.1}, in pool-based sampling approach, A learner may start to be trained with a few number of labeled instances ($L$), then request labels for one or more carefully selected unlabeled instances ($U$) using an oracle. After labeling, the new instance is simply added to the labeled set($L$), and the learner proceeds training process in a standard supervised way. This process continues up to a specified number of iterations or to achieve desired accuracy. 

\subsection{Introduction to Deep Learning}

Deep Learning is including representation learning algorithms that transform raw features to higher-level abstraction by using a deep network composed several hidden layers \citep{bengio2009learning}. In another word, deep learning applies computational approaches, which have multiple non-linear transformations to train data representation through several levels of abstraction \citep{lecun2015deep, nezhad2018predictive}. 

Deep learning applications include different areas. The most popular ones are speech detection, image recognition, automatic text generation and health informatics \citep{lecun2015deep}. In healthcare domain with explosive increase of large and high-dimensional datasets, deep learning with great performance outperformed some traditional methods in medical features representation and it showed strong potential for feature engineering and dimensionality reduction \citep{mamoshina2016applications}.

Readers for more detail about applications of deep learning in health informatics can refer to recent review papers provided by \cite{shickel2017deep}, \cite{miotto2017deep}, and \cite{ravi2017deep}.

\section{Related Works}
Deep learning and active learning as two advanced machine learning methods have been applied in different areas but there exist a few research in the literature that use the benefit of deep learning or active learning in survival analysis. In this section we review the research works which use any of those methods in survival analysis. 

\cite{vinzamuri2014active} provided the first ever active learning framework for survival analysis. They developed this approach just for regularized Cox regression survival models. Authors proposed a novel sampling strategy based on discriminative gradient for selecting the best candidate from the unlabeled pool set. Finally, they evaluated their model performance using public EHRs datasets and compared with some state of the art survival regression methods. 

In the deep learning domain, there exist few studies which developed survival analysis framework using deep learning recently. \cite{ranganath2016deep} proposed a new survival model using deep learning termed deep survival analysis. They used Deep Exponential Family (DEF) for capturing complex dependencies from clinical features including laboratory measurements, diagnosis, and medications codes. They applied their model on a large EHR dataset related to coronary heart disease. In the other research \citep{luck2017deep}, authors introduced a new deep learning approach which can directly predict the survival times for graft patients using foundations of multi-task learning. They demonstrated that their model outperforms usual survival analysis models such as cox proportional hazard model in terms of prediction quality and concordance index.

\cite{katzman2018deepsurv} proposed a cox proportional hazards deep multi-layer perceptron called DeepSurv to predict risk of event occurrence for patient and provided personalized treatment recommendations. They performed their approach on simulated and real-world datasets for testing and evaluation. Finally, They used DeepSurv on real medical studies to illustrate how it can provide treatment recommendations. In the other research, \cite{lee2018deephit} introduced a different approach called DeepHit which employs deep architecture to estimate the survival times distribution. They used neural network including two types of sub-networks: 1) a single shared sub-network and 2) family of cause-specific sub-networks. They evaluated their method based on real and synthetic datasets which illustrate that DeepHit leads to better performance in comparison with state of the art methods.  

Based on our review, there exist no study to develop a survival analysis approach using both deep learning and active learning. We address this gap in the literature to propose a deep active learning framework for survival analysis. However, There are some studies that develop deep active learning methods for other machine learning tasks. For example, \cite{zhou2013active} developed a semi-supervised learning framework termed active deep network (ADN) for sentiment analysis. They used restricted Boltzmann machines (RBM) for feature learning based on labeled reviews and large amount of unlabeled reviews, then applied gradient-descent based supervised learning for fine tuning and constructing semi-supervised framework. Finally they used active learning in their framework to improve model performance. In the other study, \cite{liu2017active} proposed a deep active learning approach using Deep Belief Network (DBN) for classifying hyperspectral images in remote sensing application. A summary of our review has been illustrated in Table \ref{summary1} which indicates no research have been developed to address a survival approach using deep learning and active learning.

\begin{table*}[t]
\begin{threeparttable}
		\footnotesize 
		\centering
		\caption {Summary of research works used deep learning or active learning in survival analysis} 
		\begin{tabular}{ p{3.4cm}  p{9cm}  c c c  }
			\hline
			\textbf{Authors} &  \textbf{Research} & \textbf{DL} &  \textbf{AL} & \textbf{SA} \\
			\hline
			\cite{zhou2013active} & proposed semi-supervised sentiment classification algorithm &  \checkmark &  \checkmark &   \\
			\hline
			\cite{vinzamuri2014active} & developed survival regression for censored
			data for electronic health records&  &  \checkmark &  \checkmark  \\
			\hline
			\cite{ranganath2016deep} & introduced a deep hierarchical generative approach for survival analysis in heart disease & \checkmark &  & \checkmark \\
			\hline
			\cite{nie20163d} & proposed a survival analysis model applied on high-dimensional  multi-modal brain images & \checkmark &  &\checkmark  \\
			\hline
			\cite{liao2016combining} & proposed a survival analysis framework using a LSTM model & \checkmark &  & \checkmark  \\
			\hline
			\cite{huang2017deep} & developed a survival model using CNN-based and one FCN-based sub-network and applied on pathological images and molecular profiles & \checkmark &  & \checkmark  \\
			\hline
			\cite{chaudhary2017deep} & introduced a DL based, survival model on hepatocellular carcinoma patients using genomic data  & \checkmark &  & \checkmark  \\
			\hline
			\cite{liu2017active} & proposed an active learning approach using DBN for classification of hyperspectral
			images   & \checkmark & \checkmark &   \\
			\hline
			\cite{luck2017deep}  & developed a patient-specific kidney graft survival model using principle of multi-task learning & \checkmark &  & \checkmark \\
			\hline
			\cite{sener2017geometric} & developed an active learning framework using CNN for image processing applications & \checkmark & \checkmark &  \\
			\hline
		    \cite{katzman2018deepsurv} & proposed a Cox proportional hazards deep neural network for personalized treatment recommendations & \checkmark &  & \checkmark  \\
			\hline
			\cite{lee2018deephit} & developed a survival model using deep learning which trained based on a loss function that uses both risks factors and survival times & \checkmark &  & \checkmark \\
			\hline
			\label{summary1}
			\vspace{0.5mm}
		\end{tabular}
	\begin{tablenotes}
		\footnotesize
		\item \textit{Note: DL, AL and SA refer to Deep Learning, Active Learning and Survival Analysis.} 
	\end{tablenotes}
		
\end{threeparttable}
\end {table*}

\section{Methodology}
The method developed in this research is an active learning based survival analysis using a novel sampling strategy. In our model, we apply deep learning for feature reduction and extraction, when data is high-dimensional, complex and sparse. Since in survival analysis we deal with censored and uncensored instances, the active learning design should be different from the regular approach. In our framework, we consider censored and uncensored instances in the training set as survival analysis needs both instances in the training process and we consider uncensored data as unlabeled instances in the pool set which their labels (time to event) are unknown.

The general framework in our survival analysis approach includes two main steps: 1) Deep feature learning for survival data and 2) Active learning based survival analysis. In the first step we do unsupervised learning using deep learning to represent features in higher level abstractions and extract data into lower dimensions. We represent both labeled (time to event) and unlabeled (censored) instances with together ($X_{train} \bigcup X_{pool}$) to obtain strong representation using pool of unlabeled data. In the other words, our framework uses the advantages of abundant unlabeled data to provide less complex and more robust features (labeled and unlabeled) for survival analysis. 
	
\begin{figure}[H]
	\centering
	\includegraphics[scale= 0.38]{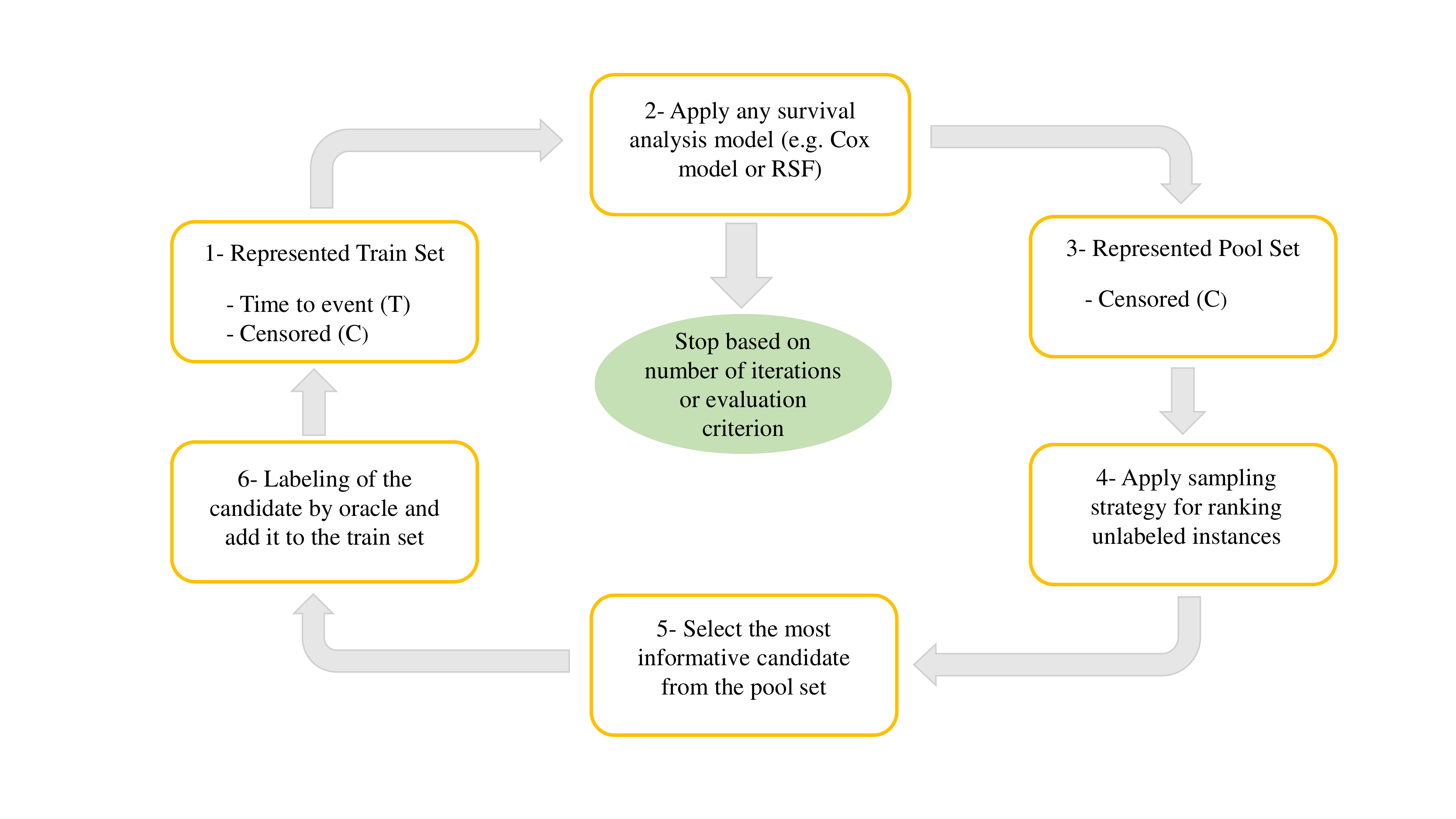} 
	\caption{Active Survival Analysis Approach}
	\label{fig.4.2}
\end{figure}

In the second step, we apply our novel active learning based survival analysis on the represented/lower dimensions features obtained from the first step. This process is demonstrated in Figure \ref{fig.4.2}.

According to this Figure, we start by applying a survival analysis method such as Cox-based regression or Random survival forest on represented train set. In the next step we use our novel sampling strategy (explained in the next section) to rank the unlabeled data based on their informativeness level. Then we select the most informative candidate from the pool and add it to the train set and repeat the process untill the stop criteria happens. 
	
\subsection{Expected Performance Improvement (EPI) Sampling (Query) Strategy}
All active learning scenarios as well as pool-based active learning use the informativeness measure for evaluation of unlabeled instances to select the best query (the most informative unlabeled instance). There exist several proposed approach which formulate such query strategies in the literature which can be categorized in general frameworks \citep{settles2010active}:1- uncertainty sampling, 2- query by committee, 3- expected model change, 4- expected error reduction, 5- variance reduction and 6- density weighted methods. 
	
In this research we developed a new sampling (query) strategy based on properties of survival analysis. In our strategy, we select the unlabeled instance as the most informative instance (the best query) when it has the greatest performance change to the current survival model if we knew its label. Our sampling model use concordance index (C-index) to define the informative measure to query the unlabeled data. The survival model is trained again by adding a new instance ($X^{+}$) from the pool to the training set: $Train_{new}$ = $Train \bigcup$ $X^{+}$ and the performance change is formulated based on the c-index difference as follows:

\begin{align}\label{equ1}
&\Delta C_{X^{+}} =\  C_{new\ model} - C_{current\ model}&
\end{align}
	
Similar to the other active learning sampling strategy, Our goal is to select the most informative instance which could maximally improve the current model performance. This selection can be formulated as follows:
\begin{align}\label{equ2}
&X^{*} = \underset{X^{+} \in pool}{\mathrm{argmax}}\ \Delta C_{X^{+}} &
\end{align}
	
Since in the real-world applications, We do not know the true label (time to event) of the instances in the pool, We should calculated the expected performance change over all possible time to events ($T_{s}$) for each unlabeled records as follows:  
\begin{align}\label{equ3}
&X^{*} = \underset{X^{+} \in pool}{\mathrm{argmax}}\ \dfrac{\sum_{s=1}^{S} h(T_{s}|X^{+}) \ \Delta C_{X^{+}}}{\sum_{s=1}^{S} h(T_{s}|X^{+})} &
\end{align}
	
Our sampling strategy works for all survival analysis approaches such as cox-based models, parametric models and random survival forests. As an example for the cox regression, $\Delta C_{X^{+}}$ can be formulated as following equation and $X^{*}$ is chosen based on Eq. \ref{equ3}.
\begin{align}\label{eq:equ4}
&\Delta C_{X^{+}}= \dfrac{1}{N} [ \sum_{\delta_{i}=1} \sum_{T_{i} <T_{j}} (\hat{\beta_{2}^{s}}X_{i} > \hat{\beta_{2}^{s}}X_{j} ) -  \sum_{\delta_{i}=1} \sum_{T_{i} <T_{j}} (\hat{\beta_{1}}X_{i} > \hat{\beta_{1}}X_{j} )] &
\end{align}
	
Where $\hat{\beta_{1}}$ and $\hat{\beta_{2}}$ are the estimated cox model coefficients trained based on the current and new training set ($Train_{new}$). $N$ refers to the comparable (permissible) pairs in validation set for calculating c-index. 
	
\subsection{Proposed Deep Active Survival Analysis (DASA) Algorithm} 
Algorithm 1 describes our deep active survival analysis approach called DASA in detail. First, we apply deep feature learning on both train and pool sets. In this step we need to keep the weights of deep network for representation learning of new instances. In line 6, we apply survival analysis on deep represented features ($Deep_{-}Survival$). This framework is flexible and all survival models can be used in this step. Then we start active learning iterations using EPI sampling strategy and update the pool and train sets until convergence. 
	
\begin{algorithm}[H]
	\caption{Deep Active Survival Analysis (DASA) Algorithm}
	\begin{algorithmic}[1]
		\small
		\Require Training set ($X_{T}$), Pool set ($X_{P}$), Survival status ($\delta$), Time to event ($T$), Deep architecture parameters (hidden layers, hidden units, ...), Active learning maximum iteration ($max_{-}iter$)  
		\State Round = 1
		\State Training deep network for feature reduction on ($X_{T} \bigcup X_{P}$)
		\State $Train$ $set\longleftarrow$ $X_{T}^{'}$
		\State $Pool$ $set\longleftarrow$ $X_{P}^{'}$
		\Repeat 
		
		\State Model = $Deep_{-}Survival$ $(X_{T}^{'}, \delta, T)$
		
		\For{each record in the pool ($x \in X_{P}^{'}$)}
		\State Apply EPI sampling strategy and calculate the expected performance improvement for each instance  
		
		\EndFor
		
		\State $X^{*}$ = $argmax_{x \in X_{P}^{'}} \dfrac{\sum_{s=1}^{S} h(T_{s}|x) \ \Delta C_{x}}{\sum_{s=1}^{S} h(T_{s}|x)} $
		\State Labeling (time-to-event) of $X^{*}$ by an Oracle based on original features 
		\State $X_{P}^{'}  \longleftarrow X_{P}^{'} -$ $\{X^{*}\}$
		\State $X_{T}^{'}  \longleftarrow X_{T}^{'} \bigcup$ $\{X^{*}\}$
		\State $ \delta_{X^{*}} \longleftarrow 1$
		\State $Round \longleftarrow Round+1$
		\Until $Round \neq max_{-}iter$
	\end{algorithmic}
\end{algorithm}
	
\subsection{Treatment Recommendations Using Proposed DASA Approach} 
In this section, we propose a simple yet effective approach to discover treatment patterns and treatment recommendations using DASA. Our method is highly useful when EHRs are high-dimensional and small size. Suppose $X_{T} =\{ X^{T}_{1}, X^{T}_{2}, ..., X^{T}_{n} \}$ is the treatment set and $X_{A} =\{ X^{A}_{1}, X^{A}_{2}, ..., X^{A}_{N} \}$ refers to all other personalized features related to each patient where $N >> n$. Therefore, the input features is the union of these two sets ($X_{T} \bigcup X_{A}$). Since in the case of high-dimensional features, traditional approaches such as cox proportional hazard or random survival forests cannot find the pattern of specific features (e.g. small treatment set), we first represent $X_{A}$ using deep learning to a lower dimension set ($X^{'}_{A}$) and then combine this represented set with the treatment set ($X_{T}$) to build the new feature set ($X_{new} = X^{'}_{A} \bigcup X_{T} $). In the second phase, we apply our active learning framework to train an accurate survival model based on new features and then find the pattern of treatment sets and interpret the results  (e.g. comparison the coefficient of treatment options using Cox model or finding the importance of different treatment plan using random survival forests).       
	
In our treatment recommendation approach, we transform many clinical features to a small feature set with higher level abstraction and more robust features. While we represent patient information to lower dimension using deep learning we combine non-represented treatment options (as features of interest) with the represented set and then perform survival analysis using active learning framework. In the next section, we demonstrate how our approach discovers the treatment patterns better than traditional approaches.

\section{Experimental Study: Survival Analysis for Prostate Cancer (SEER-Medicare Data)}
In this section, we evaluate the performance of our approach (DASA) through experimental study. We use the Surveillance, Epidemiology and End Results (SEER)-Medicare linked database from SEER program of the National Cancer Institute (NCI). SEER-Medicare data is a powerful and unique source of epidemiological data on the occurrence and survival rates of cancer in the United States. In our study, we use prostate cancer SEER-Medicare data to evaluate our survival analysis approach and provide some insights by treatment recommendation. 
	
\subsection{Datasets: SEER-Medicare Prostate Cancer Data}
	
Prostate cancer is the most popular diagnosed invasive cancer among men, with approximately 56\% of all prostate cancer patients diagnosed in men with age 65 years and older \citep{siegel2015cancer}. Fortunately, a wide range of men (nearly 90\%) are diagnosed with non-metastatic prostate cancer and 5-year relative survival rate is very high for them. The death rate for prostate cancer is different among different populations. A good example of this racial disparity is the death rate for African-American men which is 2.5 times higher than white men. there exists a critical need to develop precision survival analysis for each cohort and discover the pattern of treatment. 
	
In this study, we consider the SEER-Medicare data into two racial groups: 1) African-American patients and 2) White patients. Both groups are including many features (more than 300 features) such as demographic data, socioeconomic variables, tumor information and assigned treatment with approximately 1000 and 5000 patients respectively.    
	
Since SEER-Medicare data is high-dimensional, sparse and complex, feature representation using deep learning can build more robust features when we use pool of unlabeled data (censored instances) in the representation process. In the other hand, our method using active learning has highly potential to improve the performance of survival models when we deal with small sample size (including time-to-event and censored instances). In this way, in experimental study, we consider small samples in training of survival model and show that how our approach can improve the prediction performance in comparison with baseline. 
	
For labeling of the censored instances (unlabeled data) in active learning framework we used some scientific reports such as SEER cancer statistics review from National Cancer Institute (NCI) \citep{howlader2014seer} which acts as a prior knowledge to establish an oracle. One of these statistics is illustrated in Table \ref{t2}. 

\begin{figure}[H]
		\centering
		\captionsetup{justification=centering,margin=2.5cm}
		\includegraphics[scale= 0.5]{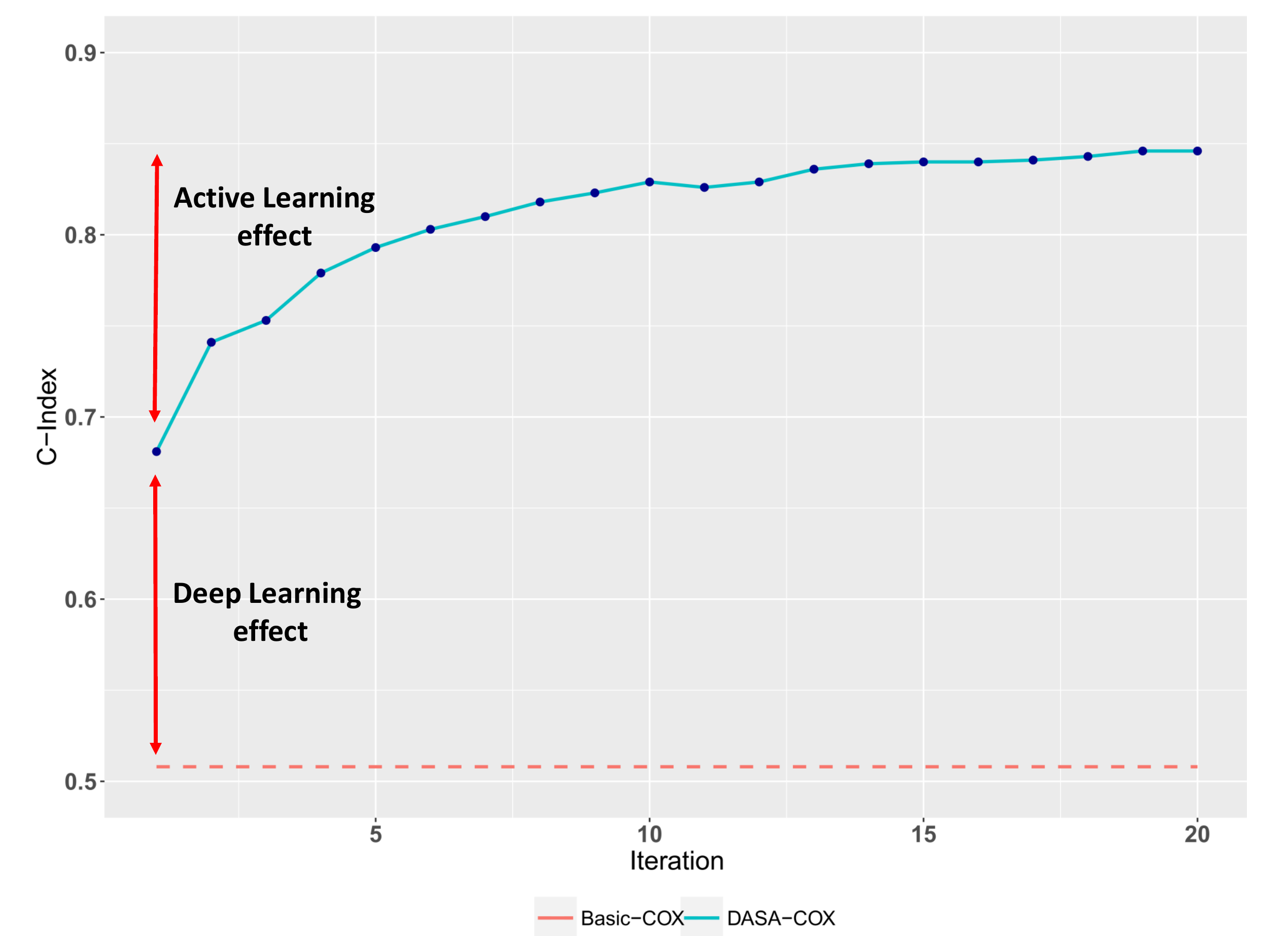} 
		\caption{Performance of proposed approach in comparison with baseline (training size =25)}
		\label{cox1}
\end{figure}

\begin{table}[H]
	\footnotesize 
	\centering
	\captionsetup{justification=justified, width=.9\linewidth}
	\caption{5-Year SEER conditional relative prostate cancer survival and 95\% confidence intervals} \label{res1}
	
	\begin{tabular}{*{4}{c}}
		\toprule
		\mc{1}{\textbf{Stage at Diagnosis}}
		&\mc{1}{\textbf{Survival Time Since Diagnosis}}
		&\mc{2}{\textbf{Percent Surviving Next 5 years}} \\
		\cmidrule{3-4}
		& &\mc{1}{Percent} & \mc{1}{Confidence Interval} \\
	    \hline 
	    
		\multirow{3}{*}{\textbf{Local}} 
		& \multicolumn{1}{c}{0-Year} & \multicolumn{1}{c}{100\%} & \multicolumn{1}{c}{(100, 100)} \\
		&\multicolumn{1}{c}{1-year} & \multicolumn{1}{c}{100\% }& \multicolumn{1}{c}{(100, 100)} \\
		&\multicolumn{1}{c}{3-year} & \multicolumn{1}{c}{100\% }& \multicolumn{1}{c}{(100, 100)} \\
		\hline
		
			\multirow{3}{*}{\textbf{Regional}} 
		& \multicolumn{1}{c}{0-Year} & \multicolumn{1}{c}{100\%} & \multicolumn{1}{c}{(100,100)} \\
		&\multicolumn{1}{c}{1-year} & \multicolumn{1}{c}{99.3\% }& \multicolumn{1}{c}{(98.9, 99.5)} \\
		&\multicolumn{1}{c}{3-year} & \multicolumn{1}{c}{98.9\% }& \multicolumn{1}{c}{(98.4, 99.2)} \\
		\hline
		
			\multirow{3}{*}{\textbf{Distant}} 
		& \multicolumn{1}{c}{0-Year} & \multicolumn{1}{c}{29.2\%} & \multicolumn{1}{c}{(28.4, 29.9)} \\
		&\multicolumn{1}{c}{1-year} & \multicolumn{1}{c}{34.1\% }& \multicolumn{1}{c}{(33.1, 35.1)} \\
		&\multicolumn{1}{c}{3-year} & \multicolumn{1}{c}{45.6\% }& \multicolumn{1}{c}{(43.9, 47.2)} \\
		\hline
		
			\multirow{3}{*}{\textbf{Unstaged}} 
		& \multicolumn{1}{c}{0-Year} & \multicolumn{1}{c}{76.6\%} & \multicolumn{1}{c}{(75.6, 77.5)} \\
		&\multicolumn{1}{c}{1-year} & \multicolumn{1}{c}{81.1\% }& \multicolumn{1}{c}{(79.8, 82.1)} \\
		&\multicolumn{1}{c}{3-year} & \multicolumn{1}{c}{82.8\% }& \multicolumn{1}{c}{(81.4, 84.1)} \\
		\hline
	
	\end{tabular}
\label{t2}
\end{table}
To evaluate the performance of our approach, we first employ CPH regression model (as a well-known survival analysis approach) and demonstrate how DASA can improve its performance based on different training sample size. For deep feature representation we used Stacked Autoencoder (SAE) deep architectures with 5 hidden layers. Figure \ref{cox1} shows the average performance of our approach for 20 iterations in comparison with baseline on the test data. We sampled training set with 25 instances from African-American patients over 10 runs and calculated the average performance in each iteration.

As demonstrated in Figure \ref{cox1}, our method (DASA-COX) improves the performance of Basic-COX significantly in terms of concordance index. This improvement is caused by two effects: 1) Deep learning effect which improve the model performance by features representation using labeled and unlabeled instances, and 2) Active learning effect which increase the model performance by involving the best labeled censored instance from the pool set in training process across all iterations.  
	
\begin{figure}[H]
		\centering
		\subfloat[African-American (50 instances) ]{\includegraphics[width=0.50\textwidth]{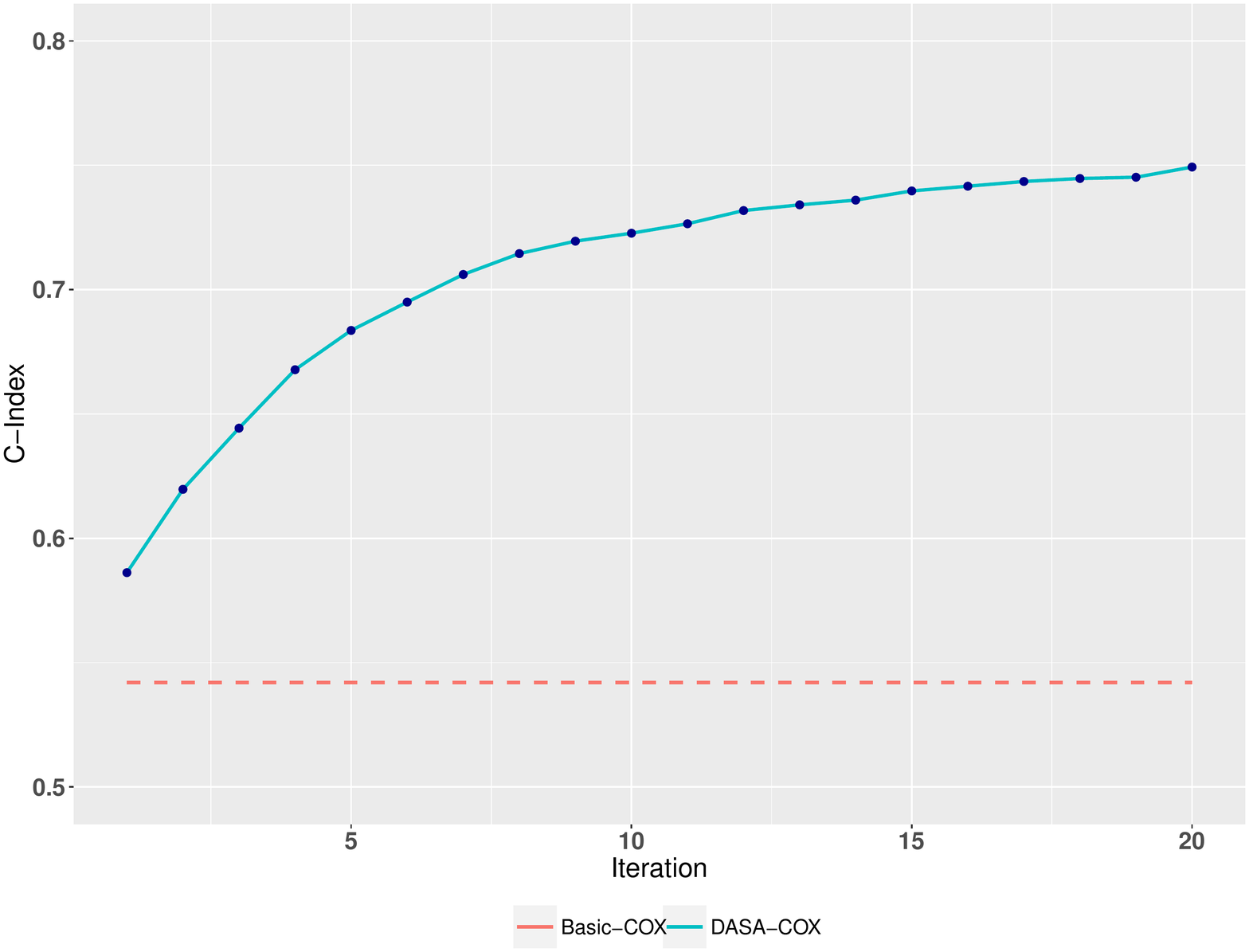}\label{fig:f1}}
		\hfill
		\subfloat[African-American (100 instances) ]{\includegraphics[width=0.50\textwidth]{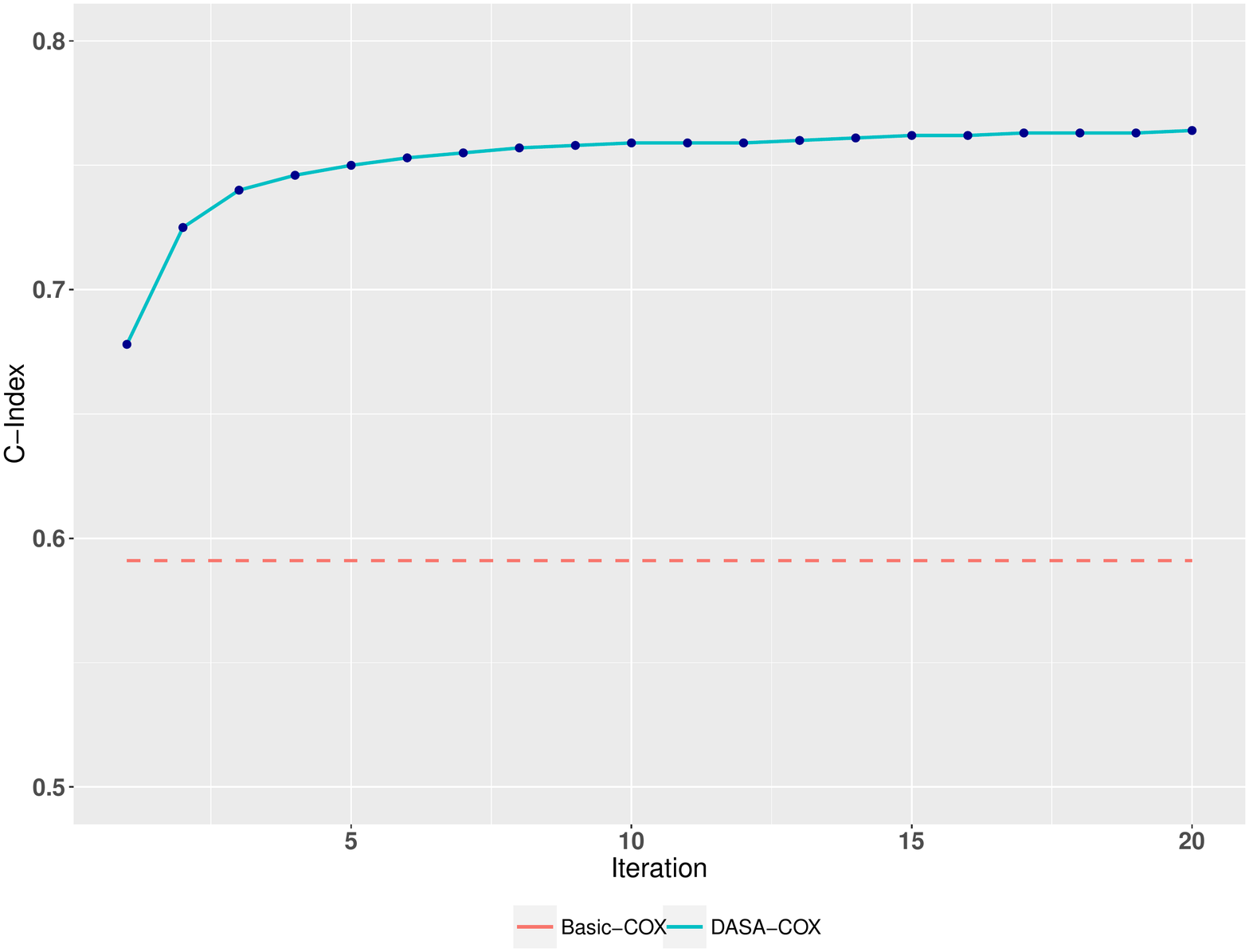}\label{fig:f2}}
		\hfill
		\subfloat[White (50 instances) ]{\includegraphics[width=0.50\textwidth]{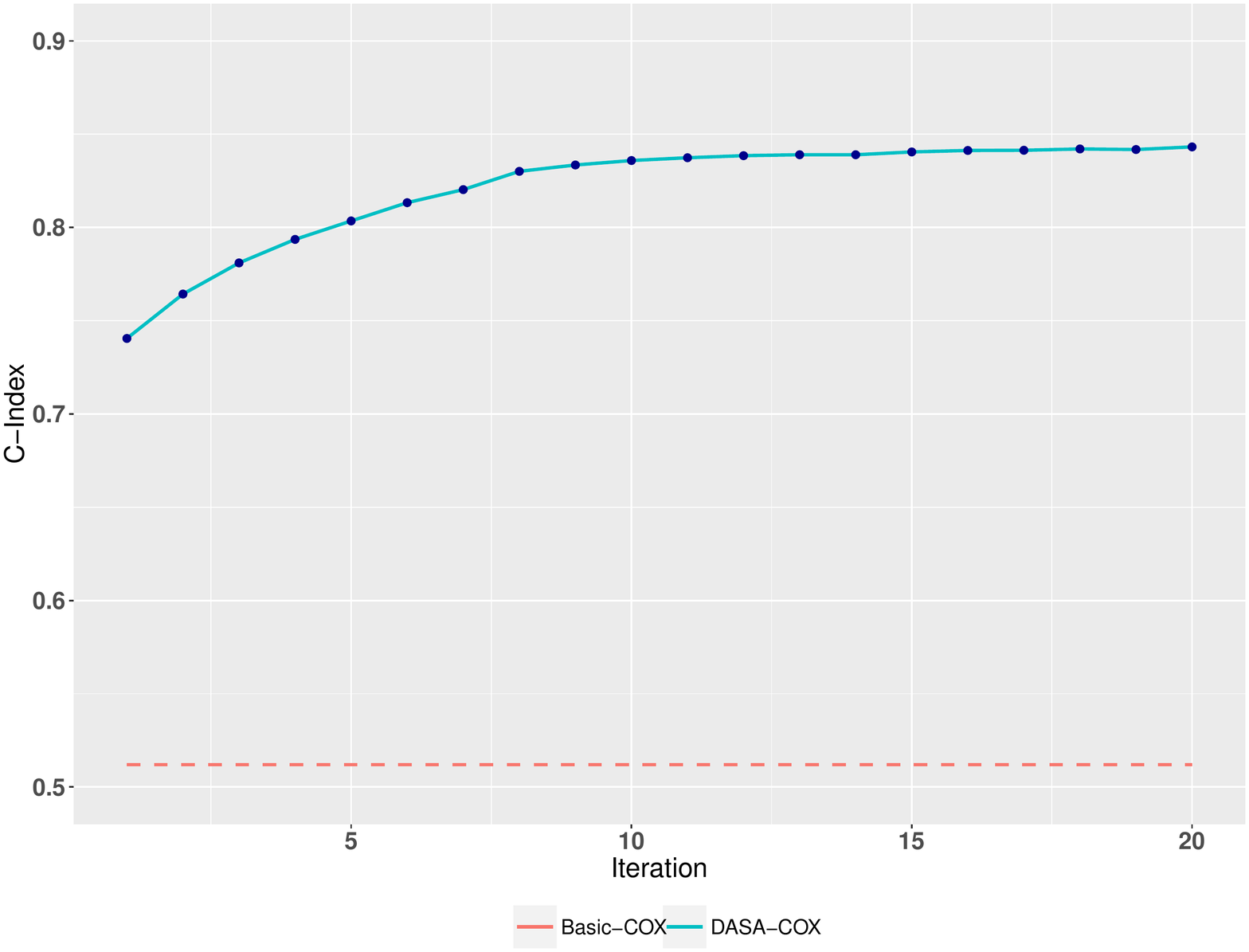}\label{fig:f3}}
		\hfill
		\subfloat[White (100 instances) ]{\includegraphics[width=0.50\textwidth]{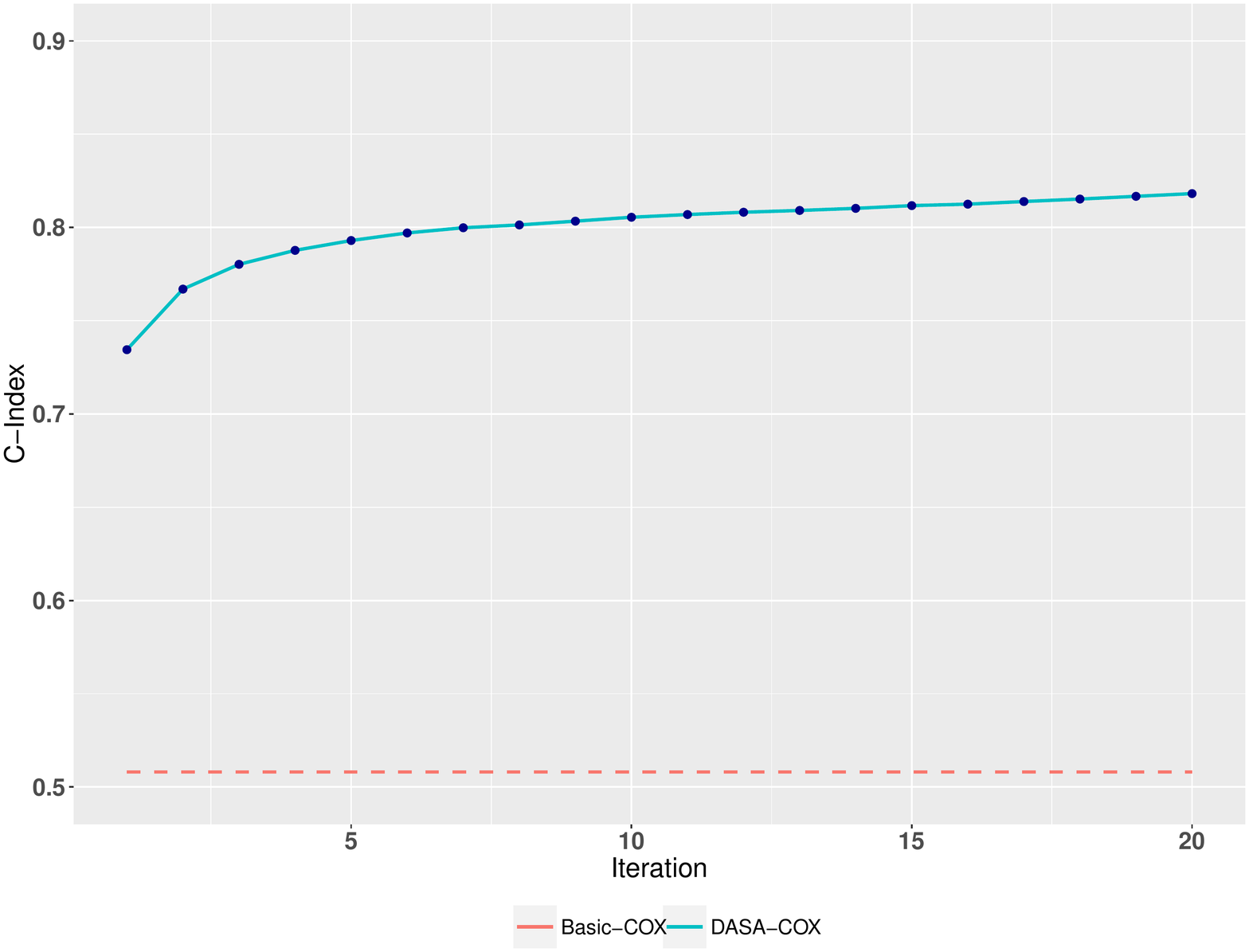}\label{fig:f3}}
		\caption{Performance of proposed approach in comparison with baseline for different training size}
		\label{cox2}
\end{figure}

Figure \ref{cox2} shows our approach performance for training size of 50 and 100 instances. Top panel belongs to African-American patients and bottom panel is related to white patients. It is clear DASA-COX outperforms baseline approach in all cases. The effect of deep learning in improving model performance is higher at the bottom panel which can be caused by larger amount of pool set related to white patients that provide better feature learning.

As mentioned before, our approach is flexible enough and can employ any survival analysis model in its framework to improve the baseline. Hence, we perform Random Survival Forests (RFS) model as a well-know non-linear survival model along with CPH model and evaluate our approach across different training sizes. The results are shown in Table \ref{res1} and \ref{res2} for African-Americans and white patients respectively.

\begin{table}[H]
		\small 
		\centering
		\captionsetup{justification=justified, width=.7\linewidth}
		\caption{Performance comparison (C-index) between DASA and baseline models (African-Americans)} \label{res1}
		
		\begin{tabular}{ |c | p{2.1cm} | p{2.2cm} | p{2.1cm} |p{2.1cm}|}
			\hline
			\textbf{Training Size} &  \textbf{CPH} & \textbf{DASA-CPH} & \textbf{RSF} &  \textbf{DASA-RSF} \\
			\hline
			\textbf{25 instances}  & 55.2\% & 84.7\% & 16.3\% & 57.6\%   \\
			\hline
			\textbf{50 instances} & 54.2\% & 74.9\% & 17.6\% & 54.5\%   \\
			\hline
			\textbf{100 instances} & 59.1\% & 76.6\% & 21.4\% & 48.2\%  \\  
			\hline
			\textbf{200 instances} & 58.6\% & 72.6\% & 22.3\% & 47.9\%  \\  
			\hline
			
		\end{tabular}
\end{table}
	
\begin{table}[H]
		\small 
		\centering
		\captionsetup{justification=justified, width=.7\linewidth}
		\caption{Performance comparison (C-index) between DASA and baseline models (Whites)} \label{res2}
		
	\begin{tabular}{ |c | p{2.1cm} | p{2.2cm} | p{2.1cm} |p{2.1cm}|}
			\hline
			\textbf{Training Size} &  \textbf{CPH} & \textbf{DASA-CPH} & \textbf{RSF} &  \textbf{DASA-RSF} \\
			\hline
			\textbf{25 instances}  & 52.4\% & 87.9\% & 13.3\% & 62.1\%   \\
			\hline
			\textbf{50 instances} & 51.2\% & 84.4\% & 15.5\% & 58.3\%   \\
			\hline
			\textbf{100 instances} & 50.8\% & 82.3\% & 15.7\% & 49.7\%  \\  
			\hline
			\textbf{200 instances} & 53.6\% & 77.1\% & 18.2\% & 46.4\%  \\  
			\hline
			
	\end{tabular}
\end{table}

The results confirm that our method can improve the concordance index significantly for cox proportional hazard model and random survival forests in each datasets. According to above results, we can conclude that DASA leads to larger performance improvement in smaller training size caused by active learning effect. 
	
In the second step, we demonstrate how our treatment recommendation approach works. we considered three well-known treatment options for prostate cancer: chemotherapy, radiotherapy and surgery as three binary variables in our dataset. Our goal is to discover the importance of each therapy using DASA approach for each subgroup of patients (African-Americans and white patients). Since in the experimental study CPH illustrated a great performance, we performed survival analysis using CPH. We do feature representation by deep stacked autoencoder network with 150, 100 and 5 hidden unites in encoder, decoder and latent layers respectively. We used small sample size with 50 instances in training process. Before training process, we combined chemotherapy, radiotherapy and surgery variables (features of interest) to the represented features came from deep learning performed on other features in training instances combined with unlabeled pool set and then trained the cox survival model using active learning framework with 20 iterations over all features. The results for average of exponential of coefficients (hazard ratios) over 10 runs shown in Table \ref{th1} for African-Americans and white patients:   
	
\begin{table}[H]
		\small
		\centering
		\captionsetup{justification=justified, width=0.8\linewidth}
		\caption{Average Hazard Ratio among different treatment plans} \label{s1}
		
		\begin{tabular}{ c | c c c c }
			\textbf{} & \textbf{Method} & \textbf{Chemotherapy} & \textbf{Radiotherapy} & \textbf{Surgery}\\
			\hline
			
			\multirow{2}{*}{\textbf{African-Americans}} 
			&\multicolumn{1}{l}{COX-Base} & \multicolumn{1}{c}{1} & \multicolumn{1}{c}{1}& \multicolumn{1}{c}{1} \\
			&\multicolumn{1}{l}{COX-DASA} & \multicolumn{1}{c}{0.74 }& \multicolumn{1}{c}{1.04}& \multicolumn{1}{c}{1.38} \\
			\hline	
			
			\multirow{2}{*}{\textbf{White Patients}} 
			& \multicolumn{1}{l}{COX-Base} & \multicolumn{1}{c}{1} & \multicolumn{1}{c}{1}& \multicolumn{1}{c}{1} \\
			&\multicolumn{1}{l}{COX-DASA} & \multicolumn{1}{c}{0.96}& \multicolumn{1}{c}{1.08}& \multicolumn{1}{c}{2.23 } \\
			
			\hline
			
	\end{tabular}
		\label{th1}	
\end{table}
	
As shown above, traditional CPH model could not differentiate between treatment plans where their hazard ratios are one. Since the data is high-dimensional traditional CPH leads to zero coefficients for these three treatment variables. On the other side, our approach using Cox model can discover the risk associated to each treatment. Based on our results, surgery has the highest risk in the both subgroup of patients, radiotherapy is associate with a decline in the survival rate while chemotherapy increases the survival rate with lowest risk. It is obvious that the pattern of hazard ratios among treatment plans are different between African-American and white patients. For example the risk related to surgery is significantly higher than the other two therapies in white patients (more than 2 times) while in the African-Americans the pattern is different.    
	
This experimental treatment recommendation was a simple example to show how our method works. This approach is highly useful for comparing the risk associated with new treatment in comparison with current treatment plans where the labeled data is rare and expensive.

\section{Discussion and Conclusion}
In this research, we proposed a novel survival analysis framework using deep learning and active learning called Deep Active Survival Analysis (DASA). Our approach is able to improve the survival analysis performance significantly and provides treatment recommendations. DASA is highly applicable when the labeled data is scarce and high-dimensional. Our approach encompasses two main phases: 1) deep feature learning and 2) active learning process. We do feature representation using deep learning to produce robust features from high-dimensional, sparse and complex EHRs. We used the advantage of pool of unlabeled data (censored instances) to provide better representation of labeled instances from deep learning implementation. In the active learning process, we developed a new sampling strategy specifically for survival analysis which can be used for any survival analysis models such as Cox-based approaches and random survival forests. 
	
In experimental study, we used SEER-Medicare data related to prostate cancer among African-Americans and white patients to demonstrate how our model can enhance the performance of survival analysis in comparison of traditional approach. Empirically we showed that deep learning has greater effect on survival performance improvement in the case that we have larger pool of unlabeled data and active learning effect is higher when we deal with smaller training sample size. We apply our treatment recommendation approach to find hazard ratio of three common treatment plan (chemotherapy, radiotherapy and surgery) for prostate cancer based on Cox model. While traditional CPH model fails to find the hazard ratios among high dimensional data, our approach discovers them and provides some racial treatment insights.     
	
In sum, our method leads to more accurate survival analysis for risk prediction, survival time estimation and treatment recommendation. Our approach is flexible enough to capture any survival analysis model and improve its performance. Our model can be applied on different areas especially in the case of testing and comparing the risk (impact) of new treatment (e.g. in healthcare) or new technology (e.g. in the manufacturing process) where the amount of labeled instances are small and labeling is expensive. For the future works, we will implement DASA on the other datasets and introduce some new sampling strategy with better performance.         

\section*{References}
\small
\bibliographystyle{model5-names}\biboptions{authoryear, round}
\bibliography{References}{}

\end{document}